\def\year{2022}\relax
\documentclass[letterpaper]{article} 
\usepackage{aaai22}  
\usepackage{times}  
\usepackage{helvet}  
\usepackage{courier}  
\usepackage[hyphens]{url}  
\usepackage{graphicx} 
\usepackage{natbib}  
\usepackage{caption} 
\DeclareCaptionStyle{ruled}{labelfont=normalfont,labelsep=colon,strut=off} 
\usepackage{algorithm}
\usepackage{algorithmic}

\usepackage{newfloat}
\usepackage{listings}
\floatstyle{ruled}
\newfloat{listing}{tb}{lst}{}
\floatname{listing}{Listing}

\usepackage{tikz}
\usepackage{booktabs}
\usepackage{soul}

\usepackage{amsmath}
\usepackage{multirow}
\usepackage{textcomp}
\usepackage[utf8]{inputenc}

\definecolor{aquamarine}{rgb}{0.5, 1.0, 0.83}
\definecolor{babypink}{rgb}{0.96, 0.76, 0.76}
\definecolor{flavescent}{rgb}{0.97, 0.91, 0.56}
\definecolor{blue}{rgb}{0.6, 0.8, 1.0}
\definecolor{purple}{rgb}{0.8, 0.4, 1.0}
\definecolor{grey}{rgb}{0.87, 0.87, 0.87}
\definecolor{orange}{rgb}{1.0, 0.8, 0.6}
\definecolor{babyblue}{rgb}{0.4, 1.0, 1.0}

\DeclareRobustCommand{\hlaquamarine}[1]{{\sethlcolor{aquamarine}\hl{#1}}}
\DeclareRobustCommand{\hlbabypink}[1]{{\sethlcolor{babypink}\hl{#1}}}
\DeclareRobustCommand{\hlflavescent}[1]{{\sethlcolor{flavescent}\hl{#1}}}
\DeclareRobustCommand{\hlblue}[1]{{\sethlcolor{blue}\hl{#1}}}
\DeclareRobustCommand{\hlpurple}[1]{{\sethlcolor{purple}\hl{#1}}}
\DeclareRobustCommand{\hlgrey}[1]{{\sethlcolor{grey}\hl{#1}}}
\DeclareRobustCommand{\hlorange}[1]{{\sethlcolor{orange}\hl{#1}}}

\DeclareRobustCommand{\hlaquamarinewithlabel}[2]{$\underbrace{\text{\hlaquamarine{#1}}}_{\text{#2}}$}
\DeclareRobustCommand{\hlbabypinkwithlabel}[2]{$\underbrace{\text{\hlbabypink{#1}}}_{\text{#2}}$}
\DeclareRobustCommand{\hlflavescentwithlabel}[2]{$\underbrace{\text{\hlflavescent{#1}}}_{\text{#2}}$}
\DeclareRobustCommand{\hlbluewithlabel}[2]{$\underbrace{\text{\hlblue{#1}}}_{\text{#2}}$}
\DeclareRobustCommand{\hlpurplewithlabel}[2]{$\underbrace{\text{\hlpurple{#1}}}_{\text{#2}}$}
\DeclareRobustCommand{\hlgreywithlabel}[2]{$\underbrace{\text{\hlgrey{#1}}}_{\text{#2}}$}
\DeclareRobustCommand{\hlorangewithlabel}[2]{$\underbrace{\text{\hlorange{#1}}}_{\text{#2}}$}

\title{Augmenting Task-Oriented Dialogue Systems with Relation Extraction}
\author{
    Andrew Lee,\textsuperscript{\rm 1}
    Zhenguo Chen,\textsuperscript{\rm 2}
    Kevin Leach,\textsuperscript{\rm 3}
    Jonathan Kummerfeld \textsuperscript{\rm 1}    
}
\affiliations{
    \textsuperscript{\rm 1} University of Michigan\\
    \textsuperscript{\rm 2} Bloomberg\\
    \textsuperscript{\rm 3} Vanderbilt University\\

    ajyl@umich.edu, Zhenguo.Chen@colorado.edu, 
	kevin.leach@vanderbilt.edu, jkummerf@umich.edu
}

\begin{document}

\maketitle

\begin{abstract}
The standard task-oriented dialogue pipeline uses intent classification and slot-filling to interpret user utterances.
While this approach can handle a wide range of queries, it does not extract the information needed to handle more complex queries that contain relationships between slots.
We propose integration of relation extraction into this pipeline as an effective way to expand the capabilities of dialogue systems.
We evaluate our approach by using an internal dataset with slot and relation annotations spanning three domains.
Finally, we show how slot-filling annotation schemes can be simplified once the expressive power of relation annotations is available, reducing the number of slots while still capturing the user's intended meaning.
\end{abstract}

\section{Introduction} \label{sec:intro}

\begin{figure}[t]
\centering\includegraphics[width=0.99\columnwidth]{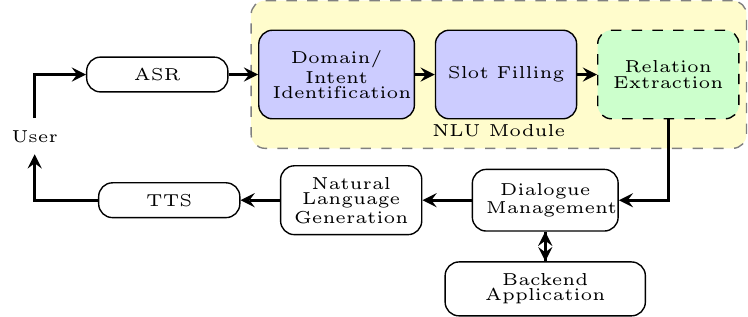}
\caption{\label{fig:dialogue-system}
The standard architecture of deployed dialogue systems.
We show how adding relation extraction can enable support for more complex queries.
}
\end{figure}

Dialogue platforms like Watson, Rasa, and Dialogflow have enabled a dramatic increase in the development and use of dialogue systems by bringing together NLP models, data collection and curation, and scalable deployment~\citep{meteer-etal-2019-tools}.
These platforms all follow the same general framework for interpreting queries, shown in Figure~\ref{fig:dialogue-system}, consisting of (1) classification models for domain or intent identification and (2) slot-filling or entity recognition models to identify relevant entities in a query.
While these two models can handle a wide range of queries, they are unable to handle more complex queries \citep{aghajanyan2020conversational, davidson2019dependency, gupta2018semantic}.
Specifically, when a query contains multiple slots with semantic \emph{relations}, such as those shown in Table~\ref{tab:example_complex_queries}, intent classification and slot-filling cannot easily capture the necessary information. 

\begin{table*}
\def\arraystretch{1.8}
\resizebox{\textwidth}{!}{
\begin{tabular}{ll}
\toprule
\textbf{Domain}                         & \textbf{Example Queries}                                                \\ \midrule
\multirow{2}{*}{Food-Order}     & Give me \hlaquamarinewithlabel{three}{quantity} \hlbabypinkwithlabel{large}{size} \hlbluewithlabel{burgers}{plus\_item} and \hlaquamarinewithlabel{two}{quantity} \hlbluewithlabel{fries}{plus\_item}.                                                           \\ 
                               & Can I get a \hlbluewithlabel{burrito bowl}{plus\_item} with \hlbluewithlabel{brown rice}{plus\_item} and \hlbluewithlabel{black beans}{plus\_item}, extra \hlbluewithlabel{chicken}{plus\_item} and no \hlflavescentwithlabel{tomatoes}{minus\_item}?      \\ \midrule
\multirow{2}{*}{Gaming}        & I'd like to see your \hlaquamarinewithlabel{fire}{enchantment} \hlbabypinkwithlabel{swords}{item} and \hlbabypinkwithlabel{shields}{item}.                                                  \\ 
                               & I'd like to see your \hlaquamarinewithlabel{fire}{enchantment} \hlbabypinkwithlabel{swords}{item} and a \hlbabypinkwithlabel{shield}{item}.             \\ \midrule
\multirow{2}{*}{Stocks} &  Show me all the \hlflavescentwithlabel{healthcare}{sector} companies in \hlaquamarinewithlabel{Europe}{location\_inside} outside of \hlbabypinkwithlabel{Germany}{location\_outside}. \\ 
                               & Which companies have a \hlbluewithlabel{2018}{date} \hlpurplewithlabel{market cap}{metric\_name} over a \hlgreywithlabel{million}{filter\_amount\_above} dollars and \hlbluewithlabel{2019}{date} \hlpurplewithlabel{revenue}{metric\_name} less than \hlorangewithlabel{2 million}{filter\_amount\_below}?                            \\ \bottomrule
\end{tabular}}
\begin{tikzpicture}[overlay]
\tikzset{bx/.style={inner sep=0pt}}
\tikzset{ba/.style={thick, ->, >=stealth}}
\node at (0,0) (invis) {};
\node[bx] at (3.9, 6.7)  (three1)   {};
\node[bx] at (4.6, 6.7)  (large1)   {};
\node[bx] at (5.6, 6.7)  (burgers1) {};
\node[bx] at (7.1, 6.7)  (two1)     {};
\node[bx] at (8.0, 6.7)  (fries1)   {};

\draw[ba] (three1.north) |- ++(.2, .215) -| (burgers1.north);
\draw[ba] (large1.north) |- ++(.2, .17) -| (burgers1.north);
\draw[ba] (two1.north) |- ++(.2, .2) -| (fries1.north);

\node[bx] at (4.6,5.6)  (bowl2)     {};
\node[bx] at (6.9,5.6)  (brownrice2) {};
\node[bx] at (9.0,5.6)   (blackbeans2){};
\node[bx] at (11.2,5.6)  (chicken2)   {};
\node[bx] at (13.4,5.6)  (tomatoes2)  {};

\draw[ba] (brownrice2.north) |- ++(-.2, .2) -| ++(bowl2);
\draw[ba] (blackbeans2.north) |- ++(-.2, .2) -| ++(bowl2);
\draw[ba] (chicken2.north) |- ++(-.2, .2) -| ++(bowl2);
\draw[ba] (tomatoes2.north) |- ++(-.2, .2) -| ++(bowl2);

\node[bx] at (5.4,4.3)  (fire3)   {};
\node[bx] at (6.6,4.3)  (swords3) {};
\node[bx] at (8.2,4.3)  (shield3) {};

\draw[ba] (fire3.north) |- ++(.2, .17) -| (swords3.north);
\draw[ba] (fire3.north) |- ++(.2, .215) -| (shield3.north);

\node[bx] at (5.4,3.25)  (fire4)   {};
\node[bx] at (6.6,3.25)  (sword4)  {};
\node[bx] at (7.3,3.25)  (shield4) {};

\draw[ba] (fire4.north) |- ++(.2, .175) -| (sword4.north);

\node[bx] at (4.6,1.8)  (healthcare5) {};
\node[bx] at (7.6,1.8)  (europe5)     {};
\node[bx] at (10.2,1.8) (germany5)    {};


\node[bx] at (6.1,.95)  (date6)       {};
\node[bx] at (7.4,.95)  (marketcap6)  {};
\node[bx] at (9.8,.95)  (million6)    {};
\node[bx] at (12.8,.95) (date26)      {};
\node[bx] at (13.8,.95) (revenue6)    {};
\node[bx] at (16.5,.95) (twomillion6) {};

\draw[ba] (date6) |- ++(.2, .2) -| ++(.7,-.2);
\draw[ba] (million6) |- ++(-.2, .2) -| ++(-2.1,-.2);

\draw[ba] (date26) |- ++(.2, .2) -| ++(.5,-.2);
\draw[ba] (twomillion6) |- ++(-.2, .2) -| ++(-2.4,-.2);

\end{tikzpicture}
  \caption{\label{tab:example_complex_queries}
  Examples of challenging utterances across three different domains that contain relations between slots.
  The output of classification and slot-filling models is not sufficient to correctly handle the examples above, and without incorporating a statistical relation extraction model, developers of a dialogue system need a workaround such as the one we describe in Section~\ref{sec:current_approach_wo_re}.
  }
\end{table*}

Meanwhile, \emph{Relation Extraction} (RE) has been effectively applied to a range of data sources, such as news articles, encyclopedia entries, and blog posts, and even conversational data \citep{yu2020dialogue}.
Recently, conversational semantic parsing \citep{aghajanyan2020conversational, cheng2020conversational, andreas2020task} has also built large end-to-end models to extract semantic relations.
While these representations are expressive, they have not been integrated into dialogue system platforms because of the difficulty of collecting new domain-specific data.

In this paper, we show how relation extraction can be incorporated into task-oriented dialogue systems.
RE is a lightweight but powerful way to extend the capabilities of a dialogue system.
In particular, suitable data can be rapidly annotated without extensive training, making domain-specific development feasible.

To evaluate our idea, we use an internal conversational dataset spanning three domains containing multiple slots and relations.
We investigate the accuracy, scalability, and generalizability of a relation extraction model within a dialogue system.
Finally we demonstrate the benefits of including an RE model in terms of the dialogue system's expressive power, meaning the system's ability to derive new semantic information that was not seen during training.
Adding relation extraction to the task-oriented dialogue pipeline will broaden the range of queries these systems can handle and improve their robustness to variation in the ways people express relations in their queries.

\section{Related Work} \label{sec:related_work}

We give a brief overview in various ways in which semantic relations can be extracted in a dialogue setting.

\subsection{Relation Extraction} \label{sec:related_work_relation_extraction}

Relation Extraction (RE) models predict labeled links between spans of text.
They have been incorporated into a range of downstream applications, including question-answering, ontology population, and knowledge base construction.
These have involved applying RE at a range of scales: sentence-level, bag-level, document-level, and few-shot \citep{han2019opennre}.
Our task is a sentence-level task, considering one dialogue utterance at a time.

Prior work has shown a variety of techniques for learning to do the task, from hand-crafting rules \citep{hearst-1992-automatic}, to supervised learning \citep{wang-etal-2016-relation, miwa-bansal-2016-end}, and weak or distant learning \citep{mintz2009distant, hoffmann-etal-2011-knowledge, zeng2015distant, lin-etal-2016-neural}.
Modeling approaches have also varied, with neural networks dominating recently \citep{goswami2020unsupervised, christopoulou-etal-2021-distantly, zhang-etal-2019-long}.
While there is an abundance of choices for models, to demonstrate the benefit of incorporating \emph{any} RE model, we consider two neural approaches, an LSTM model and a transformer.

\subsection{Task-oriented Dialogue Systems} \label{sec:related_work_dialog_systems}

A range of architectures have been explored for task-oriented dialogue systems.
Work on modular structures such TRINDI \citep{10.1017/S1351324900002539} and DIPPER \citep{bos-etal-2003-dipper} proposed ways to connect a variety of models with a structured representation of information states.
At the other end of the spectrum, end-to-end neural approaches have been explored \citep{peng2020soloist, ham2020end, hosseini2020simple}.
However, most commercial conversational platforms such as Dialog Flow, Watson, or Lex employ a modular, constrained, approach as shown in Figure~\ref{fig:dialogue-system}.
We explore an extension of this approach that can maintain the necessary accuracy level while expanding the range of supported queries.

While we focus on relation extraction, there are many other forms of structured annotation that have been considered for dialogue.
These include hierarchical slots \citep{gupta2018semantic}, dependency parses \citep{davidson2019dependency}, and abstract meaning representation \citep{bonial-etal-2020-dialogue}, such as the Alexa Meaning Representation Language (AMRL) \citep{kollar2018alexa}.
However, labeled dependency accuracy is only 78\%, while the AMR annotations are for human-robot commands and AMRL obliges to a strict ontology.
While \citet{gupta2018semantic}'s data is the closest to ours, they also take an end-to-end approach for extracting slots and intents. 
Finally, there has been work on systems that are domain general, such as the TRIPS parser \citep{aaai18trips}.
These have been applied to biomedical text and blocks world experiments, but not to the task-oriented setting we consider, and doing so would require substantial work to map the domain-general ontology to a specific need.

\subsection{Conversational Semantic Parsing}

Recently, researchers have turned to applying semantic parsing to build dialogue representations \citep{aghajanyan2020conversational, cheng2020conversational, andreas2020task}.
Unlike modular systems that represent user utterances using intent and slot information, conversational semantic parsing builds programmic representations directly from dialogue.
The resulting compositional representations allow researchers to tackle utterances that contain complex relational information.
While these representations become much richer, building and collecting programmic annotations of dialogue, which can sometimes contain application specific APIs, also becomes harder.
Our work strives to achieve a step towards the best of both worlds -- a compositional yet simple representation of user utterances.

\section{Limitations of Current Systems} \label{sec:limitations}

While the combination of intent classification and slot-filling can handle a wide range of queries, they do not always provide enough information to correctly understand a user's query.
Table~\ref{tab:example_complex_queries} includes examples of complex queries with \emph{relations} that the NLU module must be able to recognize to correctly interpret the query.
For instance, consider the two example utterances for the gaming domain.
The utterances share the same intents and slots, but the addition of the word `a' and the change from plural to singular changes the meaning.
In the first, the user is asking for ``fire swords'' and ``fire shields'', while in the second they want ``fire swords'' and any type of ``shield''.
As the table shows, relations can capture this difference, by having an edge ``fire $\leftrightarrow$ shields'' in the first, but not the second.
Not understanding the distinction in this example may lead to a fairly benign error, but the same problem appears in all applications of dialogue systems.
Even with effective intent and slot detection, without accurate relation extraction, the system may incorrectly execute a task.

\subsection{Current Approach} \label{sec:current_approach_wo_re}

In most modular dialogue systems, a user's query is passed through intent classification and slot-filling models, followed by back-end applications or external APIs, such as querying a database, aggregating results, or executing an order.
Given such a system, without a statistical relation extraction model, developers are left with writing rules in the back-end application to capture and process relations between slots. 

Within the back-end module, developers can implement heuristics to infer the relations between slots.
For instance, a developer could use the slot type, slot values, or slot position to determine the relations (e.g., in a food ordering system, if a numeric slot is immediately followed by a food item slot then make that number specify the quantity of the food item).
However, this approach has several shortcomings:

\paragraph{Accuracy.} The information captured by classifier and slot-filling models may not be sufficient to infer slot relations.
Table~\ref{tab:example_complex_queries} shows several cases where this is the case, as discussed earlier.
Handling these subtle cases would require complex rules in the back-end application, which would not be robust, causing errors in other cases.

\paragraph{Scalability.}  Heuristic relation extraction requires custom development per use case---updating the slots in the slot-filling model would require re-design and re-implementation. This makes it hard to scale such heuristics to multiple tasks or domains.

\paragraph{Generalizability.}  The developer must exhaustively consider and develop all possible slot combinations and relations.
This is not generalizable, and complex rules and code would become difficult to maintain as the number of slots and relationships grows.

These shortcomings limit the scope of dialogue systems.
As a result, developers on conversational platforms build simpler systems that only handle straightforward queries.
Incorporating relation extraction breaks this limitation, enabling support for more complex queries.

\section{Relation Extraction Module in a Dialogue System Framework}
\label{sec:re_in_dialog_sys}

We integrate relation extraction into a dialogue system by running it \emph{after} the classification and slot-filling steps.
The RE model takes their predictions as input and predicts relations in the user's utterance.
Once the RE model predicts the relations among the slots, all of the extracted information, intent, slots and relations, are passed to the back-end application.

We consider several approaches to RE, from rule-based methods to neural models.
While there may be more sophisticated models for RE, we simply use an attention-BiLSTM model and a transformer to demonstrate that incorporating \emph{any} RE model allows us to handle more complex queries.

\paragraph{Neural.} We consider two simple approaches that have been applied to relation extraction in other domains: an attention-BiLSTM model \citep{zhou-etal-2016-attention} and a transformer model \citep{alt2019improving}.
Both have been the state-of-the-art at one point on the popular SemEval 2010 Task 8 dataset \citep{hendrickx-etal-2010-semeval}.
For slot-filling we use a BiLSTM \citep{mesnil2014using}.

Like many sentence-level relation extraction approaches \citep{liu2018multiple}, these two models support classifying the correct relation between two entities within an utterance. 
However, in a conversational setting, a single utterance can have multiple slots (see Table~\ref{tab:example_complex_queries}).
We use these models to enumerate all the pairs of slots in the query utterance, and treat each slot pair as an independent relation extraction task.
For a given slot pair, we annotate the slots within the raw text utterance by adding special tokens ($\mathtt{BEGIN\_SLOT}$, $\mathtt{END\_SLOT}$) as well as the slot's label before and after the slot tokens.
This enumeration step is not required for models that can identify the correct relation among multiple slots, such as \citet{liu2018multiple}.

\paragraph{Rule-Based.} We also consider a baseline that is indicative of the approach used when there is no machine learning based relation extraction model in the NLU pipeline (Section~\ref{sec:current_approach_wo_re}).
This heuristic was designed to be generic enough to be applicable across all three of the domains we consider.
The only domain-specific information is a set of rules in which each slot is identified as a \emph{modifier} and/or \emph{modified} slot (or neither).
For each \emph{modifier} slot type, we identify a list of valid \emph{modified} slot types and a relation type the slot pair entails.
Using this list, we iterate over slots that appear in a query utterance.
For each \emph{modifier} slot $x$, we determine which (if any) of the other slots $Y$ are valid \emph{modified} slots, according to the configured rules.
If a single \emph{modified} slot $y \in Y$ exists, we assign the specified relation between $x$ and $y$.
When multiple $y_i \in Y$ exist, we select the slot in $Y$ that appears nearest to $x$ in the utterance.

\section{Evaluation} \label{sec:evaluation}

In this section, we describe our internal dataset and how we measure the performance of our proposed approach.

\subsection{Conversational Data with Relation Annotation} \label{sec:data}

The dataset that was used for our evaluation consists of utterances repurposed from slot-filling data in a production dialogue system.
The relations amongst slots were annotated manually using a custom annotation tool. 
Details about the dataset, including the number of utterances, the domains, slots, and relations are described in our supplementary material.

For some of the domains, extra crowdsourced data or new slot types were added to introduce more examples of relations.
In all cases, the relations were manually annotated.

\paragraph{Stocks.}  This data was originally collected for a heuristic RE approach (Section~\ref{sec:current_approach_wo_re}).
As a result, the types of queries handled in the task were intentionally chosen to be solvable without relation extraction.
This makes it a relatively difficult case for our approach to show improvements.
There are two versions of this dataset: one with the original slot labeling scheme with no relation annotation, and one with the same utterances but a new slot labeling scheme as well as relation annotations (described in Section~\ref{sec:scenario_2}).

\paragraph{Food-Order.} These utterances were repurposed from an existing slot-filling dataset, with added crowdsourced data to include more utterances with relations and to provide a challenging benchmark.
When collecting new data, we provided workers with examples and asked them to write paraphrases, then replaced slot values with new options from a list of food menu items to add diversity to the slots.
To ensure the utterances were diverse we sampled them in a two step process.
First, we grouped utterances based on the pattern of slots they contain (e.g., [ QUANT, PLUS ] for ``Give me \{ QUANT 3 \} \{ PLUS burgers \}'').
Next, we randomly selected a set of slot patterns and randomly picked a set of utterances for each pattern.

\paragraph{Gaming.} This dataset was built similarly to the food-ordering dataset, but with extra slots and example utterances added prior to crowdsourcing.

\subsection{Results} \label{sec:scenario1-analysis}

We seek to develop RE models that are highly accurate, scale well to complex utterances, and generalize well to sentence structures not seen in training.

\begin{figure*}[t]
\centering\includegraphics[width=0.95\textwidth]{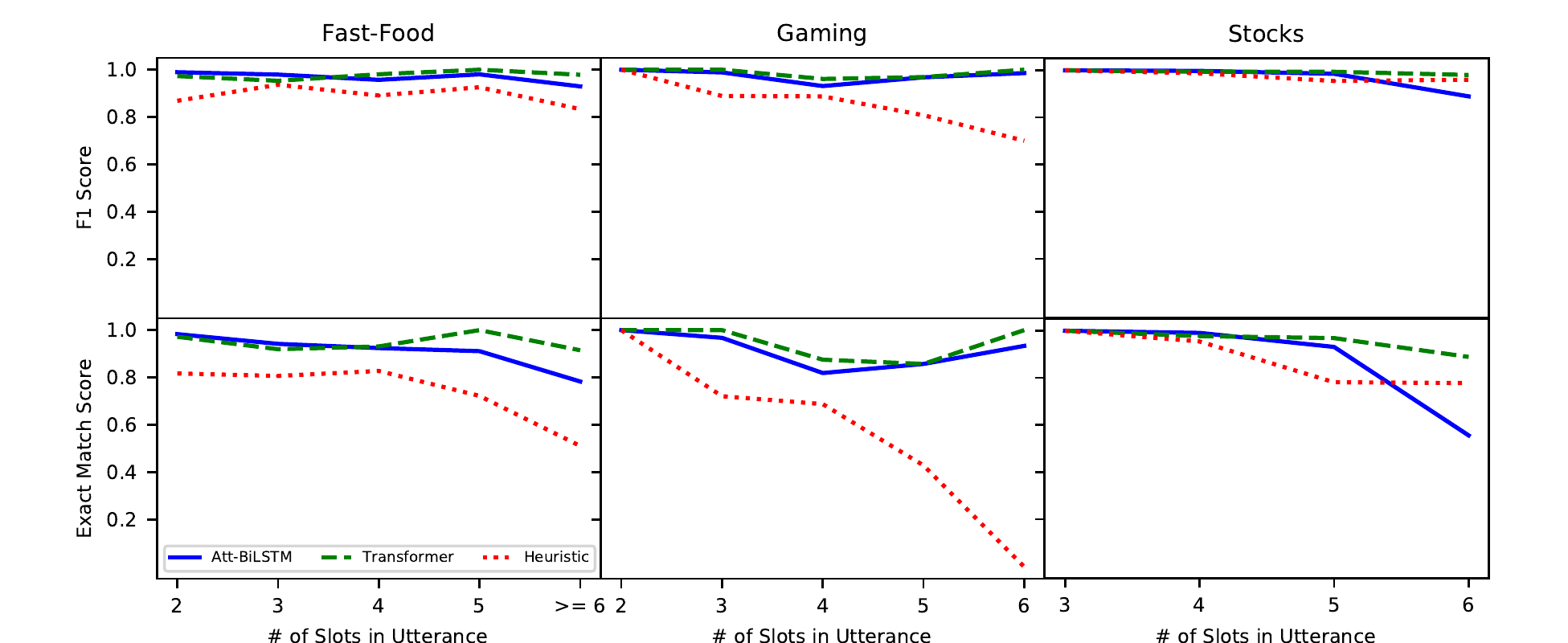}
\caption{\label{fig-scalability}
The $F_1$ and exact match scores across RE approaches versus the number of slots in an utterance. The performance of the heuristic approach does not scale well as the queries become more complex. Note that in the gaming domain for the 5 and 6 slot cases there are only 7 and 3 utterances in the test set respectively.
}
\end{figure*}

\begin{table}
\resizebox{\columnwidth}{!}{
    \begin{tabular}{l rr | rr | rr}
    \toprule
                     & \multicolumn{2}{c}{Food}        & \multicolumn{2}{c}{Gaming}          & \multicolumn{2}{c}{Stocks} \\ 
                     & $F_1$   & EM & $F_1$  & EM & $F_1$ & EM  \\
    \midrule
    \multicolumn{1}{l}{Heuristic}               & 88    & 74    & 88     & 75     & 98    & 94   \\
    \multicolumn{1}{l}{Att-BiLSTM}         & 96    & 92    & 97 & 93 & 98 & 97  \\ 
    \multicolumn{1}{l}{Transformer}          & \textbf{97} & \textbf{94} & \textbf{98}    & \textbf{96}     & \textbf{99}    & \textbf{98}  \\ 
    \bottomrule
    \end{tabular}}
\caption{\label{tab:generic-f1-scores}
The $F_1$ and exact match scores for each RE approach for three domains.
The exact match score is the percentage of utterances in which all relations were correctly identified.
There are 227, 114, and 189 test utterances for each domain.
Bold indicates the highest score in each column.
}
\end{table} 

\begin{figure}[t]
\centering\includegraphics[width=0.95\columnwidth]{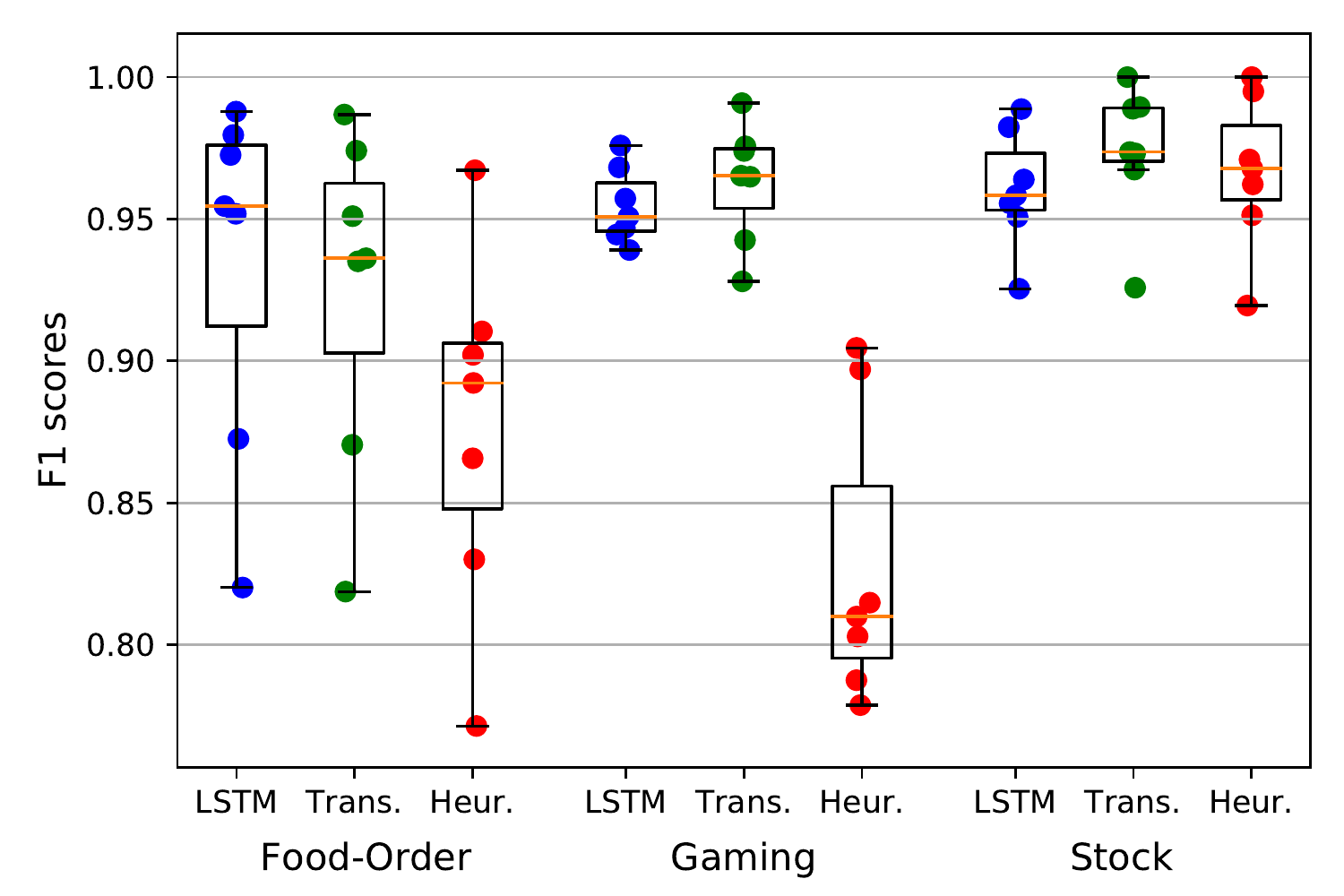}

\caption{\label{fig-slot-pattern}
$F_1$ score of RE approaches when trained and tested on different slot patterns (the order in which slots appear in a query utterance).
These scores indicate how well a model performs on slot patterns, and thus a sentence structure, that it has not seen during training.
}
\end{figure}

To measure these characteristics, we evaluate the $F_1$ score and \emph{exact match score} across several experiments.
The exact match score indicates the percentage of utterances in which every relation in the utterance is correctly predicted.
These scores measure the RE models' predictions in isolation (they are provided with oracle slot-filling and intent classification output).
For each neural network, we used the hyperparameters specified in the original work across all experiments (we verified they were effective using a development set in each domain).
We also use the development dataset for early stopping during training.

\begin{table*}
\def\arraystretch{1.8}

\resizebox{\textwidth}{!}{
\begin{tabular}{l|l}
\toprule
    \addlinespace[-\aboverulesep] 
\multicolumn{2}{l}{\textbf{Labeling Schemes}}                                             \\ \midrule
Slot-Based &   Show me all the companies in \hlaquamarinewithlabel{Europe}{location\_inside} outside of \hlbabypinkwithlabel{Germany}{location\_outside}. \\ 
Relation-Based & Show me all the companies in \hlbluewithlabel{Europe}{location} \hlflavescentwithlabel{outside}{negation\_modifier} of \hlbluewithlabel{Germany}{location}. \\ \midrule 
Slot-Based &     Show me the \hlaquamarinewithlabel{EBITDA}{query\_metric} of companies that have a \hlbabypinkwithlabel{market cap}{filter\_metric} over a \hlpurplewithlabel{million}{filter\_amount\_above} dollars and \hlbabypinkwithlabel{revenue}{filter\_metric} less than \hlorangewithlabel{2 million}{filter\_amount\_below}?  \\ 

Relation-Based & Show me the \hlbluewithlabel{EBITDA}{metric} of companies that have a \hlbluewithlabel{market cap}{metric} \hlflavescentwithlabel{over}{filter\_modifier} a \hlgreywithlabel{million}{amount} dollars and \hlbluewithlabel{revenue}{metric} \hlflavescentwithlabel{less}{filter\_modifier} than \hlgreywithlabel{2 million}{amount} ?                            \\ \bottomrule
\end{tabular}}
\begin{tikzpicture}[overlay]
\tikzset{bx/.style={inner sep=0pt}}
\tikzset{ba/.style={thick, ->, >=stealth}}
\node at (0,0) (invis) {};

\node[bx] at (7.8, 3.0) (germany_table2) {};
\draw[ba] (germany_table2) |- ++(0.2, .2) -| ++ (1.5, -.2);

\node[bx] at (8.1, 0.85) (market_cap_table2) {};
\node[bx] at (10.3, 0.85) (over_table2)   {};
\draw[ba] (over_table2) |- ++(.0, .2) -| ++ (-1.3, -.2);
\draw[ba] (over_table2) |- ++(.0, .2) -| ++ (1.3, -.2);

\node[bx] at (15.05, 0.85) (less_table2) {};
\draw[ba] (less_table2) |- ++(.1, .2) -| ++ (-1.2, -.2);
\draw[ba] (less_table2) |- ++(.1, .2) -| ++ (1.7, -.2);

\end{tikzpicture}
  \caption{\label{tab:new_label_scheme}
  A standard labeling scheme for slot filling (slot-based), compared to a new labeling scheme with relations incorporated (relation-based).
  The standard slot-filling scheme requires each slot to capture both the meaning of the slot tokens as well as the meaning of their context.
  The new approach allows the labeling scheme for slot-filling to be simplified, while the relation extraction model now captures the contextual information.
  }
\end{table*}

\paragraph{Accuracy.} Table~\ref{tab:generic-f1-scores} presents $F_1$ and exact match scores for all three approaches on all three datasets, averaged over five runs.
The consistently high $F_1$ and exact match scores indicate that these models are appropriate in task-oriented dialogue systems.
Moreover, while additional developer effort could increase the heuristic performance, our approach requires no such effort.

\paragraph{Scalability.} We consider \emph{scalability} to be our ability to identify relations in increasingly complex utterances.
First, we split the test set based on the number of slots per utterance.
We then measure each model's performance on each subset of the test data.
The $F_1$ and exact match scores averaged across five runs are presented in Figure~\ref{fig-scalability}.
As the test utterances increase in complexity, the rule-based approach quickly deteriorates in performance, while the neural-network approaches maintain robust performance.

\paragraph{Generalizability.}
Lastly, we follow the query-split idea from \citet{acl18sql} to evaluate the \emph{generalizability} of the RE approaches by testing on sentence structures not seen during training.
For each dataset, we created new train-test splits based on \emph{slot patterns}: the order in which slot labels appear in an utterance.
We grouped all of the utterances by their patterns and randomly placed each group in either the train or test set.
This meant that none of the patterns of slots in test set utterances are seen in training.

We repeated this experiment seven times in each domain for each RE approach.
The resulting $F_1$ scores are shown in Figure~\ref{fig-slot-pattern}.
We prefer a model that consistently generalizes well (i.e., tightly distributed $F_1$ scores).
While the neural networks seem to generalize to new slot patterns better than the heuristic approach, some of the domains have a wide range in $F_1$ scores across all three RE approaches.
We hypothesize that either (1) some slot patterns have difficult relations to identify, or (2) some slot patterns are crucial for the model to see during training to generalize well.

Note that the stocks dataset was originally crafted to perform well with a heuristic RE approach (Section~\ref{sec:data}), and thus performs similarly to the other RE approaches in all three experiments.

\section{Expanding Expressive Power Using Slots And Relations}
\label{sec:scenario_2}

Once we have a pipeline that includes a relation extraction model in addition to intent classification and slot-filling, we can revisit previous design decisions.
In particular, to get around the lack of a relation extraction model, many slot-filling schemes are designed with variations on slots that actually capture contextual information.
In this section, we explore how slot-filling schemes can be simplified without a decrease in performance, once a relation extraction model is available.

Consider the examples in Table~\ref{tab:new_label_scheme}.
The first example uses $\mathtt{location\_inside}$ and $\mathtt{location\_outside}$ slots to distinguish locations to include or exclude, while the second uses $\mathtt{filter\_amount\_above}$ and $\mathtt{filter\_amount\_below}$.
Note that these slots are capturing not only the meaning (location, amount) of the slot value, but also the context (inside versus outside, above versus below).

Instead, we propose a simplified slot-filling labeling scheme in which we eliminate contextual meanings per slot.
For example, we could simplify the slots above to $\mathtt{location}$ and $\mathtt{amount}$ slots, and then capture relations between those slots with new modifier tokens within the utterances.
By using an RE model to identify these logical relations, we can simplify the overall slot-filling model.
Moreover, we can potentially increase the model's generalizability---by learning logical modifiers and existing relationships they have within tokens, the model can predict relations between previously unseen pairs of logical modifiers and slots.

\begin{figure}[t]
\centering\includegraphics[width=0.85\columnwidth]{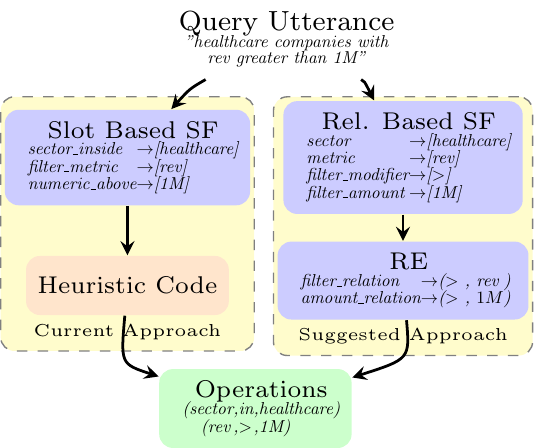}

\caption{\label{fig:scenario2_fig}
Two query pipelines with different labeling schemes for slot-filling: a slot-based scheme and a relation-based scheme.
To compare the two pipelines, we annotated each utterance as operation objects that represent the required actions to fulfill the query.
}
\end{figure}

This new labeling scheme allows us to treat slots as \emph{operands} and relations as \emph{operators}.
These operands and operators give us more expressive power by representing any utterance as a logical expression.
For instance, any $\mathtt{location}$ slot that does not have a relation with any $\mathtt{negation\_modifier}$ would be equivalent to the $\mathtt{location\_inside}$ slot, and those that do have a relation would be equivalent to the $\mathtt{location\_outside}$ slot.

This expressive power also gives the NLU module flexibility to be improved, because we no longer need training data for every possible combination of a slot and relation.
For instance, imagine a dialogue system that has been trained for a $\mathtt{location\_outside}$ slot and a $\mathtt{location\_inside}$ slot.
However, once this system is deployed in production, a user may ask about ``neighboring locations'' (``What are the top companies in the area surrounding Boston?'').
Similarly, while a dialogue system knows to recognize $\mathtt{filter\_amount\_above}$ or $\mathtt{filter\_amount\_below}$ slots, a user may ask about an amount that is ``equal to'', ``greater than or equal to'', or ``less than or equal to'' a certain value.
This requires new training utterances to be collected to cover every possible variation for each slot, and hinders the flexibility and scalability of slot-filling models.
On the other hand, if the slot-filling model is able to extract both the simplified slot and the operational tokens, and the relational models are able to extract the correct relations between new slot pairs, new semantic information can be derived.

\subsection{Results: Logical Expressions}\label{sec:scenario2-analysis}

To evaluate the approach of using slots as operands and relations as operators, we first manually annotate each utterance in the test set of the stocks domain into a set of ``operations.''
These operations represent the tasks that the back-end application needs to execute.


Figure~\ref{fig:scenario2_fig} contains examples of operations and a summary of the two query pipelines that we compare: (1) the approach without a relation extraction model, under the standard (slot-based) labeling scheme for slot-filling, and (2) our approach with a simplified (relation-based) slot-filling model, followed by a RE model.
The operation annotations allow us to compare the end-to-end performances of the two approaches (the $F_1$ scores across the three different NLU models in the two query pipelines are not directly comparable, as they are each learning different tasks).
In this case, the exact match score indicates the percentage of utterances in which all the correct operations were derived.

\begin{table}
\begin{tabular}{l r r}
    \toprule
        Method & $F_1$ & Exact Match \\
    \midrule
        Slot-based SF & 97 & 79 \\
        \multirow{2}{*}{Logical Expression} & 97 (SF) & \multirow{2}{*}{85} \\
                                            & 98 (RE) & \\
    \bottomrule
\end{tabular}
\caption{\label{tab:scenario2-accuracy}
The $F_1$ and exact match scores for the two pipelines as depicted in Figure~\ref{fig:scenario2_fig}. The exact match score indicates the end-to-end performances by measuring the percentage of utterances that derives the correct operations.
There are 189 test utterances.
}
\end{table}

\begin{table*}
\resizebox{\textwidth}{!}{
    \begin{tabular}{cc | ccccc | cccc}
\toprule
                \multicolumn{2}{c}{Data Size}        & \multicolumn{5}{c}{Overall Accuracy}          & \multicolumn{4}{c}{Per Slot or Relation $F_1$} \\ 
                  &   & \multicolumn{2}{c}{Slot-Filling $F_1$} & & \multicolumn{2}{c|}{End-to-end EM} &
                  s.b.~SF & r.b.~SF & r.b.~SF & RE \\
                Train  &  Test & s.b. & r.b. & RE & s.b. & r.b. &
                  [Sector\_Out] & [Sector] & [Neg\_Mod] & [Neg\_Rel]  \\
                  \midrule
                 0  & 90  & 53 & 89 & 81 & \phantom{\ \ \ }0    & \phantom{\ } 4  & \phantom{\ } 0    & 78 & 97 & 51  \\ 
                 8  & 90  & 58 & 92 & 91 & \phantom{\ \ }8  & 33 & 10 & 85 & 98 & 81  \\ 
                 16 & 90  & 58 & 93 & 96 & 12 & 59 & 17 & 88 & 97 & 95  \\ 
                 32 & 90  & 74 & 91 & 96 & 40 & 57 & 66 & 85 & 98 & 94  \\ 
                 64 & 90  & 85 & 93 & 98 & 67 & 66 & 81 & 88 & 99 & 98  \\ 
    \bottomrule
    \end{tabular}}
\caption{\label{tab:scenario2-scalability-sp}
Accuracy comparison ($F_1$ and exact match scores) of the two query pipelines (see Figure~\ref{fig:scenario2_fig}) on a test set that contains a new slot (\texttt{sector\_outside}) that was never seen during training.
s.b. (slot-based) SF and s.b. E2E indicate the slot-filling and end-to-end performance of the slot-based query pipeline.
r.b. (relation-based) SF, RE, and r.b. E2E indicate the slot-filling, RE, and end-to-end performance of the new query pipeline.
The difference between the s.b. and r.b. approaches is in how the utterances are annotated (see Table ~\ref{tab:new_label_scheme}).
The attention-BiLSTM was used for the RE model in this table. Transformer accuracy is elided, but yielded similar results.
}
\end{table*} 

\begin{table}
\centering
\begin{tabular}{cc  ccc}
        \toprule
            \multicolumn{2}{c}{Data Size}    & \multicolumn{2}{c}{$F_1$ Scores}  \\ 
            Train  & Test  & AttBiLSTM & Transformer  \\
        \midrule
             0  & 50  & 60 & 58  \\ 
             8  & 50  & 90 & 84  \\ 
             16 & 50  & 87 & 91  \\ 
             32 & 50  & 94 & 97  \\ 
             64 & 50  & 95 & 98  \\ 
        \bottomrule
\end{tabular}
\caption{\label{tab:scenario2-scalability-gdc}
$F_1$ scores on a relation ($\mathtt{enchantment}$) seen between a new pair of slots.
The RE models originally only sees examples of the $\mathtt{enchantment}$ relation between $\mathtt{enchantment} \leftrightarrow \mathtt{item}$ slots during training, but are tested on the same relation between a new slot pair, $\mathtt{enchantment} \leftrightarrow \mathtt{monster}$.
As training examples with this relation in the new slot pair are introduced, the models converge to high $F_1$ scores quickly.
}
\end{table} 

\paragraph{Accuracy.} Table~\ref{tab:scenario2-accuracy} presents the $F_1$ scores and the exact match scores of the two approaches.
The use of logical expression has a higher end-to-end performance, and given the similarity in performance between the two slot-filling models, the improvements are most likely coming from the high performance of the RE model.

\paragraph{Scalability.} In this case we consider \emph{scalability} to be the system's ability to be flexible by increasing its expressive power.
We demonstrate the scalability of our approach by building a new test set in the stocks domain in which all examples contain a new slot $\mathtt{sector\_outside}$ that never occurs in the training data.
However, the training data contains two similar slots, one with the same slot type ($\mathtt{sector}$) and another that involves negation ($\mathtt{location\_outside}$).
This allows the slot-filling model to recognize the $\mathtt{sector}$ slot and for the RE model to capture the contextual meaning ($\mathtt{negation\_relation}$), in turn allowing the whole system to derive the previously-unseen semantic information ($\mathtt{sector\_outside}$).

Table~\ref{tab:scenario2-scalability-sp} presents $F_1$ and exact match scores of the RE model and the slot-filling models under the different labeling schemes.
For $F_1$, we also show scores for the specific relations and slots.
We also capture how the performance scales as we introduce more training data that contains $\mathtt{sector\_outside}$.
These accuracy numbers indicate that the combination of slots and relations allow us to derive the semantic information for a new slot that was not seen during training.
In contrast, the heuristic approach has no way to derive a slot that it has not seen during training (and must see enough training data to learn about a new slot).

\paragraph{Generalizability.} Table~\ref{tab:scenario2-scalability-gdc} similarly demonstrates that a RE model can generalize and predict the correct relation between slot pairs that were not seen during training.
In this experiment, we created a new train and test split of the gaming domain dataset, such that during training, the model sees examples of the $\mathtt{enchantment}$ relation between the $\mathtt{enchantment} \leftrightarrow \mathtt{item}$ slot pairs, but never between the $\mathtt{enchantment} \leftrightarrow \mathtt{monster}$ slot pairs.
In contrast, the new test data contains the $\mathtt{enchantment}$ relation between the $\mathtt{enchantment} \leftrightarrow \mathtt{monster}$ slot pair in every utterance.
The $F_1$ scores with zero additional training data indicate that the models can still identify the correct relation between slot pairs that were never seen with a relation during training.
They also converge with high accuracy with small amounts of additional training data.
Similar results are shown in Table~\ref{tab:scenario2-scalability-sp} for the $\mathtt{Negation}$ relation for the relation extraction model (far right column).

\section{Conclusion} \label{sec:conclusion}

In this paper we propose the \emph{addition} of a relation extraction model to the NLU pipeline in modular task-oriented dialogue systems.
As part of evaluating the use of relation extraction, we manually annotated an internal dataset from a production dialogue system across 3 different domains that include slot and relation annotations.
We examine the benefits of incorporating a relation extraction model by presenting a set of evaluations that cover accuracy, scalability, and generalizability.
We further introduce and evaluate the notion of building logical expressions with the use of slots and relations, which can lead to more expressive power, allowing the NLU module to be more flexible and scalable to future changes.


\bibliography{aaai22.bib} 
\end{document}


\maketitle

\appendix

\section{Data} \label{sec:appendix}

In the following section we include details about the datasets used in this work.
The sizes of all three domains of the dataset are presented in Table~\ref{tab:dataset-sizes}.

\begin{table}[H]
    \centering
    \begin{tabular}{l|lll}
        Domain     & Train & Dev & Test  \\ \toprule
        Food-Order &  1059 & 227 & 227 \\
        Gaming     &  529  & 114 & 114 \\
        Stocks     &  882  & 189 & 189
    \end{tabular}
    \caption{\label{tab:dataset-sizes}
    Each column shows the number of utterances in the train ($70\%$), development ($15\%$), and test ($15\%$) set of each domain.}
\end{table}

We also list all of the relations and their corresponding slot pairs per domain in Table~\ref{tab:food-order-relations}, Table~\ref{tab:gaming-relations}, and Table~\ref{tab:stock-relations}.
The relations in this work are non-directional.
Any slot pair without a relation specified by default has a ~\texttt{None} relation.
A slot can only have a single relation with another slot, but can have a relation with multiple slots.

\begin{table}[H]
\centering
    \begin{tabular}{ll|l}
        \multicolumn{3}{c}{Food-Order} \\ \toprule
        \multicolumn{2}{c}{Slots} & \multicolumn{1}{c}{Relation} \\ \toprule
        plus  & plus     & add\_topping \\
        plus  & minus    & remove\_topping    \\
        plus  & quantity & numeric         \\
        plus  & size     & size            \\
        minus & quantity & numeric        \\
        minus & size     & size        
    \end{tabular}
    \caption{\label{tab:food-order-relations}
    The relations and their slot pairs in the food-order domain.}
\end{table}

\begin{table}[H]
    \centering
    \begin{tabular}{ll|l}
        \multicolumn{3}{c}{Gaming} \\ \toprule
        \multicolumn{2}{c}{Slots} & \multicolumn{1}{c}{Relation} \\ \toprule
        item        & size    & size        \\
        item        & cost    & cost        \\
        map         & monster & loation     \\
        enchantment & item    & enchantment \\
        enchantment & monster & enchantment
    \end{tabular}
    \caption{\label{tab:gaming-relations}
    The relations and their slot pairs in the gaming domain.}
\end{table}

\begin{table}[H]
\resizebox{\columnwidth}{!}{
    \begin{tabular}{ll|l}
        \multicolumn{3}{c}{Stock} \\ \toprule
        \multicolumn{2}{c}{Slots} & \multicolumn{1}{c}{Relation} \\ \toprule
        filter\_modifier & metric\_name       & filter\_metric\_relation \\
        filter\_modifier & amount             & filter\_amount\_relation \\
        location         & negation\_modifier & negation\_relation       \\
        date\_metric     & metric\_name       & date\_relation           \\
        sector\_name     & negation\_modifier & negation\_relation
    \end{tabular}}
    \caption{\label{tab:stock-relations}
    The relations and their slot pairs in the stocks domain.}
\end{table}